\documentclass[letterpaper, 10 pt, conference]{ieeeconf}  
\IEEEoverridecommandlockouts                              % This command is only needed if 
\usepackage{graphicx}

\usepackage{tabularx}
\usepackage{soul} %to cross out the font
\usepackage{comment}
\usepackage[font=footnotesize]{caption}
\usepackage[font=footnotesize]{subcaption}

\usepackage[utf8]{inputenc}
\usepackage[export]{adjustbox}
\usepackage{wrapfig}
\usepackage{array}
\usepackage{multirow}
\usepackage{booktabs}
\usepackage{float}
\usepackage{cite} %this responsible for the reference in the line [12]-[14] not [12],[13],[14] 

\usepackage{dsfont}
\usepackage{url}
\usepackage{mathtools} %matrix alignment
\usepackage{xcolor}

\usepackage{comment}

\usepackage{lipsum}% http://ctan.org/pkg/lipsum

\usepackage{amsmath}
\usepackage{bm}
\newcommand{\bA}{\bm{A}}
\newcommand{\bM}{\bm{M}}
\newcommand{\bC}{\bm{C}}
\newcommand{\bg}{\bm{g}}
\newcommand{\bq}{\bm{q}}
\newcommand{\btau}{\bm{\tau}}
\newcommand{\bu}{\bm{u}}
\newcommand{\bF}{\bm{F}}
\newcommand{\bT}{\bm{T}}
\newcommand{\bJ}{\bm{J}}
\newcommand{\bV}{\bm{V}}
\newcommand{\bw}{\bm{w}}
\newcommand{\bQ}{\bm{Q}}
\newcommand{\bP}{\bm{P}}
\newcommand{\bG}{\bm{G}}
\newcommand{\bR}{\bm{R}}
\newcommand{\bX}{\bm{X}}
\newcommand{\bXi}{\bm{\Xi}}
\newcommand{\bY}{\bm{Y}}
\newcommand{\bU}{\bm{U}}
\newcommand{\bzero}{\bm{0}}
\usepackage{xcolor}

\newtheorem{thm}{Theorem}

\graphicspath{ {images/} }
\title{\LARGE \bf
Optimal Oscillation Damping Control of\\cable-Suspended Aerial Manipulator with a Single IMU Sensor
}
\author{Yuri S. Sarkisov$^{1,2}$, Min Jun Kim$^{1}$, Andre Coelho$^{1}$, \\Dzmitry Tsetserukou$^{2}$, Christian Ott$^{1}$, and Konstantin Kondak$^{1}$ % <-this % stops a space
	\thanks{The funding of the European Commission to the AEROARMS project under the H2020 Programme (Grant Agreement 644271) is acknowledged.}% <-this % stops a space
	\thanks{$^{1}$The authors are with Institute of Robotics and Mechatronics, German Aerospace Center (DLR), Wessling, Germany.}
	\thanks{$^{2}$The authors are with Space CREI, Skolkovo Institute of Science and Technology (Skoltech), Moscow, Russia.}
	\thanks{
	{\tt\footnotesize e-mails: iurii.sarkisov@dlr.de, minjun.kim@dlr.de, andre.coelho@dlr.de, d.tsetserukou@skoltech.ru, Christian.Ott@dlr.de, Konstantin.Kondak@dlr.de}}%
	}

\begin{document}
\maketitle
\thispagestyle{empty}
\pagestyle{empty}

%%%%%%%%%%%%%%%%%%%%%%%%%%%%%%%%%%%%%%%%%%%%%%%%%%%%%%%%%%%%%%%%%%%%%%%%%%%%%%%%
\begin{abstract}
This paper presents a design of oscillation damping control for the cable-Suspended Aerial Manipulator (SAM). The SAM is modeled as a double pendulum, and it can generate a body wrench as a control action. The main challenge is the fact that there is only one onboard IMU sensor which does not provide full information on the system state. To overcome this difficulty, we design a controller motivated by a simplified SAM model. The proposed controller is very simple yet robust to model uncertainties. Moreover, we propose a gain tuning rule by formulating the proposed controller in the form of output feedback linear quadratic regulation problem. Consequently, it is possible to quickly dampen oscillations with minimal energy consumption. The proposed approach is validated through simulations and experiments.
\end{abstract}

%\begin{keywords}
%\DB{Aerial Systems: Applications} 
%\end{keywords}

%%%%%%%%%%%%%%%%%%%%%%%%%%%%%%%%%%%%%%%%%%%%%%%%%%%%%%%%%%%%%%%%%%%%%%%%%%%%%%%%

\section{Introduction}
\label{sec:Intro}

Aerial manipulation is a modern and prospective field in interaction robotics with a significant number of industrial applications, especially in remotely located and dangerous environment \cite{ruggiero2018aerial, khamseh2018aerial, baturone2018aeroarms, kim2018stabilizing}. Recently a new branch in this field has begun to emerge: a robotic manipulator is decoupled from the aerial carrier using, for instance, a cable \cite{suarez2018lightweight, miyazaki2019long}. The main motivation for such an approach includes the ability to operate in a narrow and complex environment.%, (iii) and reducing the influence of the carrier on the manipulation device, e.g., UAV's downwash and ground effect. 

The main challenge to utilize such systems in real world scenarios is the pendulum motion caused by the cable. It is important to damp out the oscillation as quickly as possible when it occurs due to any disturbances such as the motion of aerial carrier, robotic arm's motion, or wind gust. To this end, one may control the aerial carrier \cite{lee2015study, yoshikawa2017damping} to cancel out the oscillations. However, these methods can provide only indirect damping for the oscillations.

\begin{figure}[t]
	\centering
\includegraphics[width=0.95\linewidth]{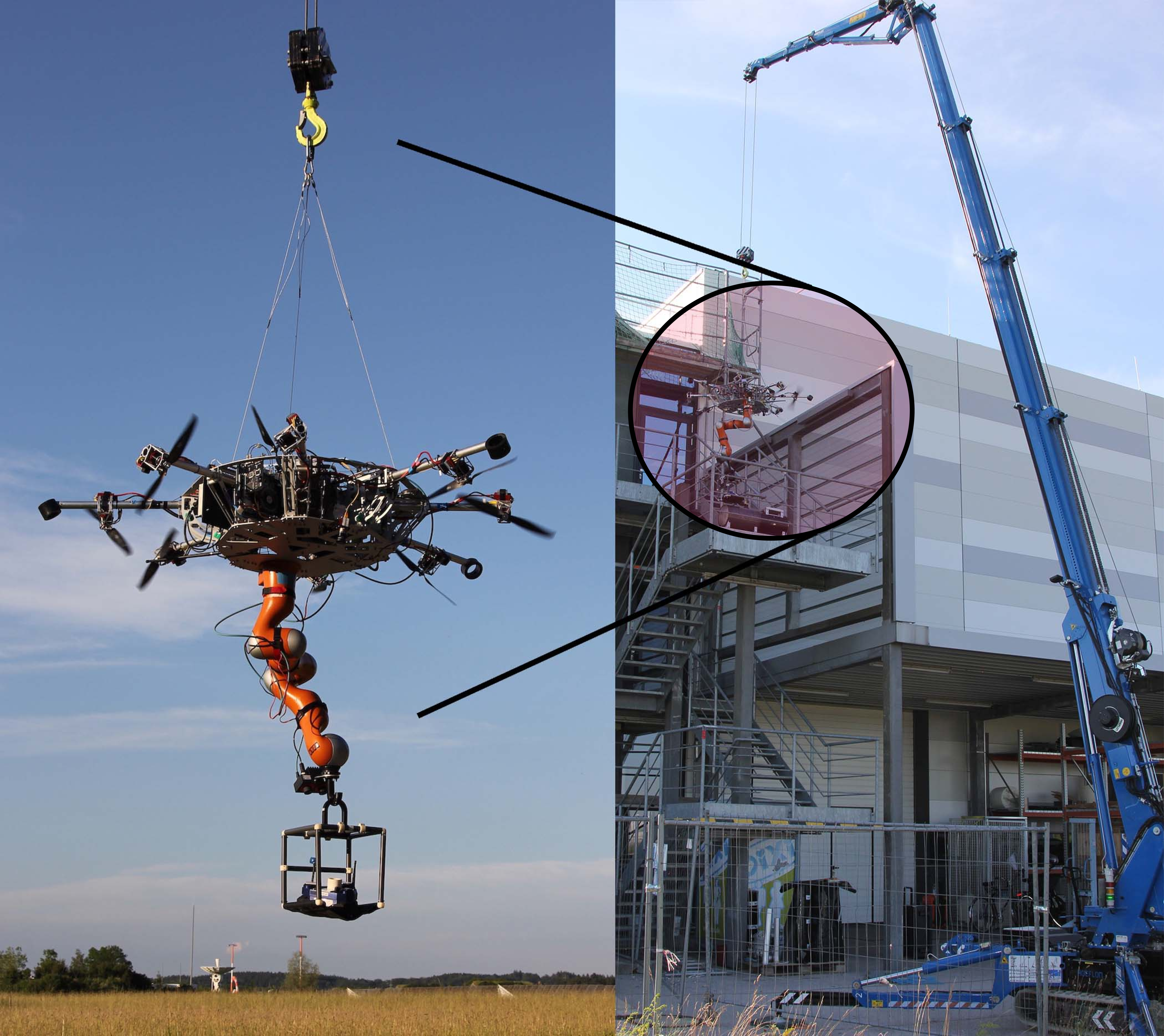} 
	\caption{The cable-suspended aerial manipulator (SAM) in action.}
	\label{fig:main}
\end{figure}

To damp out the oscillations directly, we can consider to have actuation means on the manipulation platform \cite{kim2018oscillation}. We recently developed so-called cable-Suspended Aerial Manipulator (SAM) platform which is equipped with propulsion units to dampen any oscillations, see Fig. \ref{fig:main}. Since it is suspended by a cable, the diameter of propellers might be small as its own weight is supported by the aerial carrier; one may refer to \cite{sarkisov2019development, JongSeok20, coelho20} for more details about design and application of the SAM. Therefore, the manipulator can perform an arbitrary manipulation task while the multi-rotor platform is responsible for the oscillation damping. 

%Also, one of the main features of SAM is that it is suspended by a cable, meaning that size of propellers can be reduced as its self weight is supported by the aerial carrier.

%, the second one includes systems which stabilize manipulator base locally \cite{kim2018oscillation}.

However, as will be discussed in detail later, the SAM behaves like a double pendulum, not a single pendulum. Since the motion of a double pendulum is more complex, the onboard IMU sensor does not provide all the states needed for the damping control. 

%In this paper, therefore, we propose a single IMU-based oscillation damping controller for the SAM platform. In this paper, we show only local stability for the proposed controller, but we show it is almost globally stable by numerical simulations. In fact, since the equilibrium point of interest is stable even without any controllers, the proposed controller is expected to be stable by large as it injects some damping to the system. One important thing in addition to stability is that, as addressed earlier, oscillation damping should be accomplished as quick as possible, because oscillation motion hinders precise manipulation. Moreover, 

%, which ensures a stable base of the robotic arm via the damping out oscillation around roll and pitch of the platform while keeping constant yaw. The proposed controller is based on the output-feedback linear quadratic regulation (LQR). The fundamental principle of the controller is based on the possibility to extract the information about the angular velocity of the crane's chain from the state of the platform. Design, stability analysis, selection of the optimal gains,  simulation results, and experimental validation of the developed controller are presented. 

In this paper, to overcome this issue, we design a controller motivated by a simplified model of the SAM. Moreover, we consider two criteria in the control design. First, as addressed earlier, oscillation damping should be accomplished as quickly as possible to perform manipulation tasks. Second, since the aerial system is operated by a battery, we should take the power consumption into account. To this end, we seek an optimal controller that minimizes linear quadratic function that balances two criteria. Through the simulation and experimental validations, the proposed controller turns out to be robust while having a simple form.

%In this paper, we show only local stability for the proposed controller, but we show it is almost globally stable by numerical simulations. In fact, since the equilibrium point of interest is stable even without any controllers, the proposed controller is expected to be stable by large as it injects some damping to the system. One important thing in addition to stability is that, as addressed earlier, oscillation damping should be accomplished as quick as possible, because oscillation motion hinders precise manipulation. Moreover, 

The rest of the paper is organized as follows. Section \ref{sec:Sam} gives a brief overview of the SAM platform and presents its mathematical model. Section \ref{sec:control} describes control design, stability analysis, and optimal gain selection strategy. Section \ref{sec:validation} shows simulation and experimental validation of the designed controller. Finally, section \ref{sec:end} concludes the paper.

\begin{figure}[t]
	\centering
	\includegraphics[width=1\linewidth]{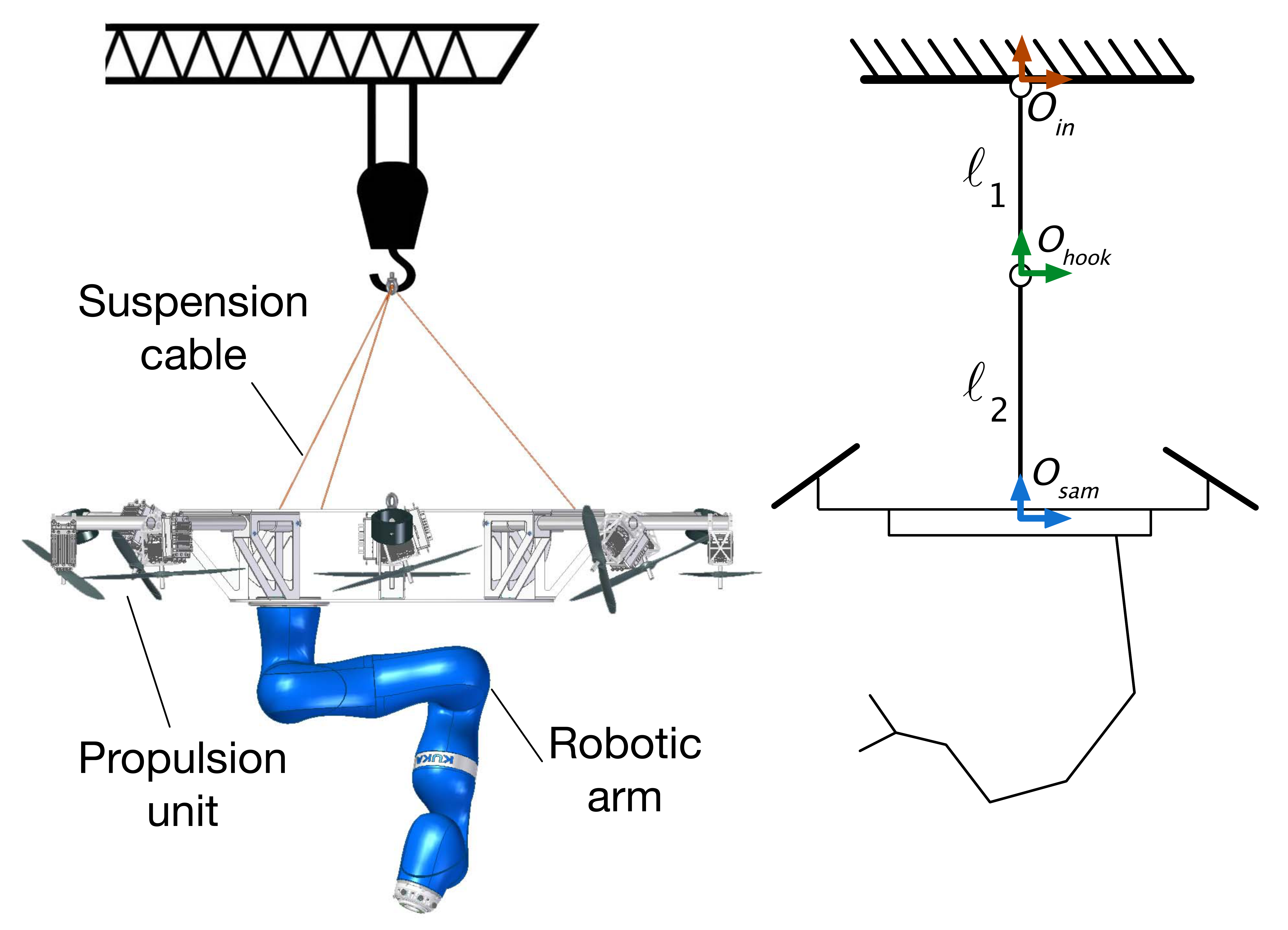}
	\caption{The SAM paltform.}
	\label{fig:platform}
\end{figure}

\section{The SAM platform}
\label{sec:Sam}
In this section, we briefly introduce the SAM platform, including actuation, sensing systems, and mathematical model.
\subsection{System overview}

\begin{figure}[t]
	\centering
	\includegraphics[width=1\linewidth]{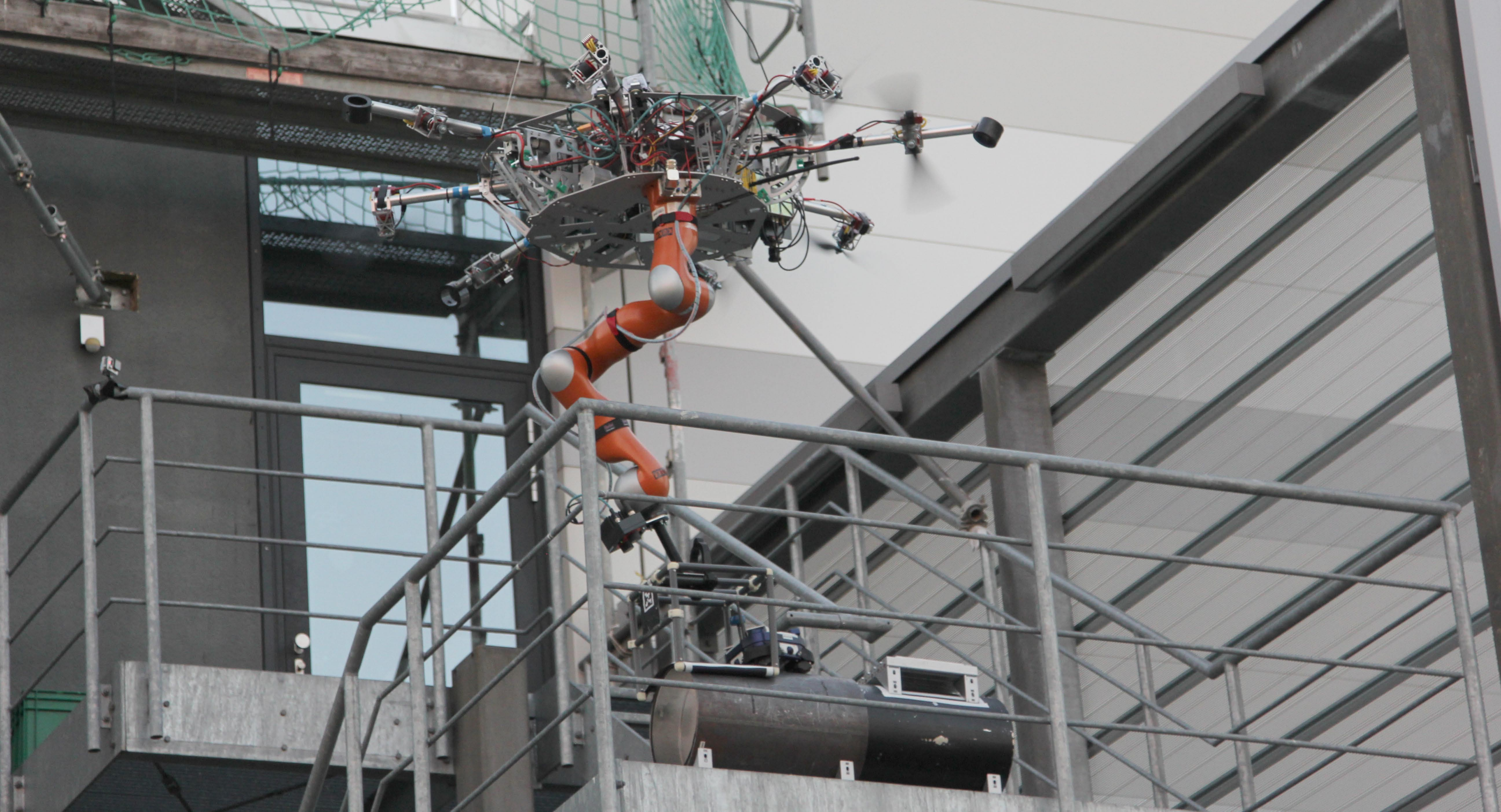}
	\caption{Crawler deployment on the pipe using the SAM platform.}
	\label{fig:industry}
\end{figure}

The SAM is an aerial manipulation platform which was designed to perform various manipulation tasks in complex industrial sites. As shown in Fig. \ref{fig:platform}, the SAM is suspended on the hook of an overhead crane by means of cables\footnote{Overhead crane can be often found in many industrial spots. Depending on the application, other carriers, e.g., mobile crane or manned/unmanned helicopter, can be exploited with the SAM platform.} and equipped with a robotic arm. In pair with the crane, the SAM can approach most of the task locations. Once it is close to the target, a torque-controlled 7 degrees of freedom (DoF) robotic arm performs a manipulation task. As an example, Fig. \ref{fig:industry}  shows how SAM is performing mobile crawler deployment on the pipe. In this task, the SAM should press the cage of the crawler at the top of the pipe with certain force to maintain enough contact force. This task is challenging because rather precise positioning should be maintained (to avoid collision) under the environmental contact which may significantly disturb SAM's motion.

To accomplish this task, self-stabilization of the SAM is essential. In particular, this paper is interested in oscillation damping of the SAM, because it is important to dissipate the SAM's motion in order to perform manipulation precisely. To this end, the SAM is equipped with 8 propulsion units with full actuation capability. By properly orienting the propellers, omnidirectional wrench (i.e., 6 DoF torques and forces) can be generated at the geometric center of the SAM platform \cite{franchi2018full}. 
%To avoid readers' confusion, we underline once again that the SAM is a fully actuated platform (6DoF force torque can be commanded). Namely, unlike the typical quadrotors, it can move horizontally without tilting, and vice versa.

The SAM is equipped with various sensors including IMU, 3D-vision camera, and GPS with RTK support. For our controller, we rely only on the IMU sensor, because it is the most robust one among all options. Indeed, GPS may not provide an accurate position in a complex industrial or indoor environment, and the reliability of the vision sensor is limited by the lighting conditions. In contrast, modern IMU with AHRS functionalities exploits real-time gyroscope drift correction and yaw adaptation to the disturbed magnetic environment \cite{xsens}.

In this paper, by exploiting the full actuation, we propose an optimal oscillation damping controller of the SAM, using a single onboard IMU sensor.% (more precisely, angular velocity and yaw angle measurements).

\subsection{Modeling of the SAM}

The SAM suspended on the crane can be modeled as a spherical double pendulum, where the first link is a chain between crane tip and crane hook with length $l_1$, and the second link is the distance between hook and platform itself with length $l_2$, see Fig. \ref{fig:platform}. Thus, the state of such a system can be described by 6 variables: the Euler angles of the first ($\bq_1 \in \Re^3$) and the second ($\bq_2 \in \Re^3$) spherical passive joints of the double pendulum.   %: $\bq_1 = [\phi_1, \theta_1, \psi_1]^T \in \Re^3$ and $\bq_2 = [\phi_2, \theta_2, \psi_2] \in \Re^3$ , i.e., the Euler angles of the first and second spherical passive joints of the double pendulum.

Let us introduce three coordinate frames. SAM frame ($O_{sam}$) is located at the center of mass (CoM) of the SAM platform, while its  $z$ axis is aligned with the second link and directed upward. Hook frame ($O_{hook}$) is placed at the second spherical passive joint, and its $z$ axis is aligned with the first link and also directed upward. Finally, $O_{in}$ represents the inertial frame. Initially, its orientation coincides with the SAM frame at the moment when onboard IMU is initialized. Onboard IMU provides orientation and angular velocity $\bw_b$ of the platform relative to the inertial frame. The weight of the links is neglected, and the two masses, which correspond to the weight of the hook and SAM, are located at the origins of the $O_{hook}$ and $O_{sam}$ frames respectively.

It is worth mentioning that joints of the double pendulum are not actuated, i.e., they are passive. To control the SAM, as described previously, we can apply body wrench $\bu=[\bF^T \; \bT^T]^T$ at the origin of $O_{sam}$ frame using propulsion units. Based on the preceding description, the equation of motion for the SAM can be written as: 
\begin{align}
\label{eq:dynamics_model_full}
\bM (\bq) \ddot{\bq} + \bC (\bq, \dot{\bq}) \dot{\bq} + \bg (\bq) = 
\begin{bmatrix}
\bm{J^Tu} \\ \bm{\tau_{m}}
\end{bmatrix}, 
\end{align}
where $\bM$ is the inertia matrix, $\bC$ is the centrifugal/Coriolis terms, and $\bg$ is the gravity vector. The configuration $\bq$ is:
\begin{align}
\label{eq:config_orig}
\bq =
[\bq_1^T \; \bq_2^T \; \bq_m^T]^T.
\end{align}
Here, $\bq_m,\btau_{m}$ represent the generalized coordinates and torque input of the robotic arm. $\bm{J}$ is the Jacobian matrix that maps $[\dot{\bq}_1^T \; \dot{\bq}_2^T]^T$ to body twist. 

\section{Oscillation Damping Control of the SAM}
\label{sec:control}
\subsection{Control goal}

The equilibrium point (or operating point) of interest corresponds to the bottom position with the stretched configuration of the SAM, i.e., where the potential energy is minimum. Any external perturbation in this position can cause oscillation of the SAM. In fact, due to the presence of the internal joint damping, the system itself is asymptotically stable to the equilibrium point without any control. However, the value of the damping is very low, so natural stabilization takes a long time which is not acceptable for real industrial scenarios. Therefore, our goal is to design an oscillation damping controller to achieve faster damping behavior with minimal power consumption. One challenge is that, as will be shown shortly, the IMU sensor does not provide enough information for the control. To overcome this, in the following section, we restrict the SAM's motion to a plane to get some useful insights.

%In the following, we design an oscillation damping controller using only onboard IMU. Moreover, we seek optimal control gains in the sense of linear quadratic function.

\subsection{Reduced model for control design}
\label{sec:reduced_model}

In this paper, we consider a decentralized control approach. Namely, the SAM control and manipulator control are decoupled. Regardless of chosen control strategy \cite{albu2007unified, tsetserukou2008vibration} and compensators \cite{iskandar2019dynamic, kim2019model} for robotic arm, from SAM's point of view, the dynamic behavior of manipulator is treated as an external disturbance that causes oscillations.
% The purpose of oscillation damping control is to damp out any motion caused by the disturbance. 
Also, experimentally we find out that yaw control of the SAM can be independently performed because (i) we have a good control authority in yaw direction, and (ii) dynamics of yaw is rather decoupled from the others around the operating point. Indeed, we could achieve very strong yaw control by applying a common geometric control approach \cite{lee2010geometric}.

\begin{figure}[t]
	\centering
    \begin{subfigure}[]{0.47\linewidth}
		\centering
    	\includegraphics[width=1\linewidth]{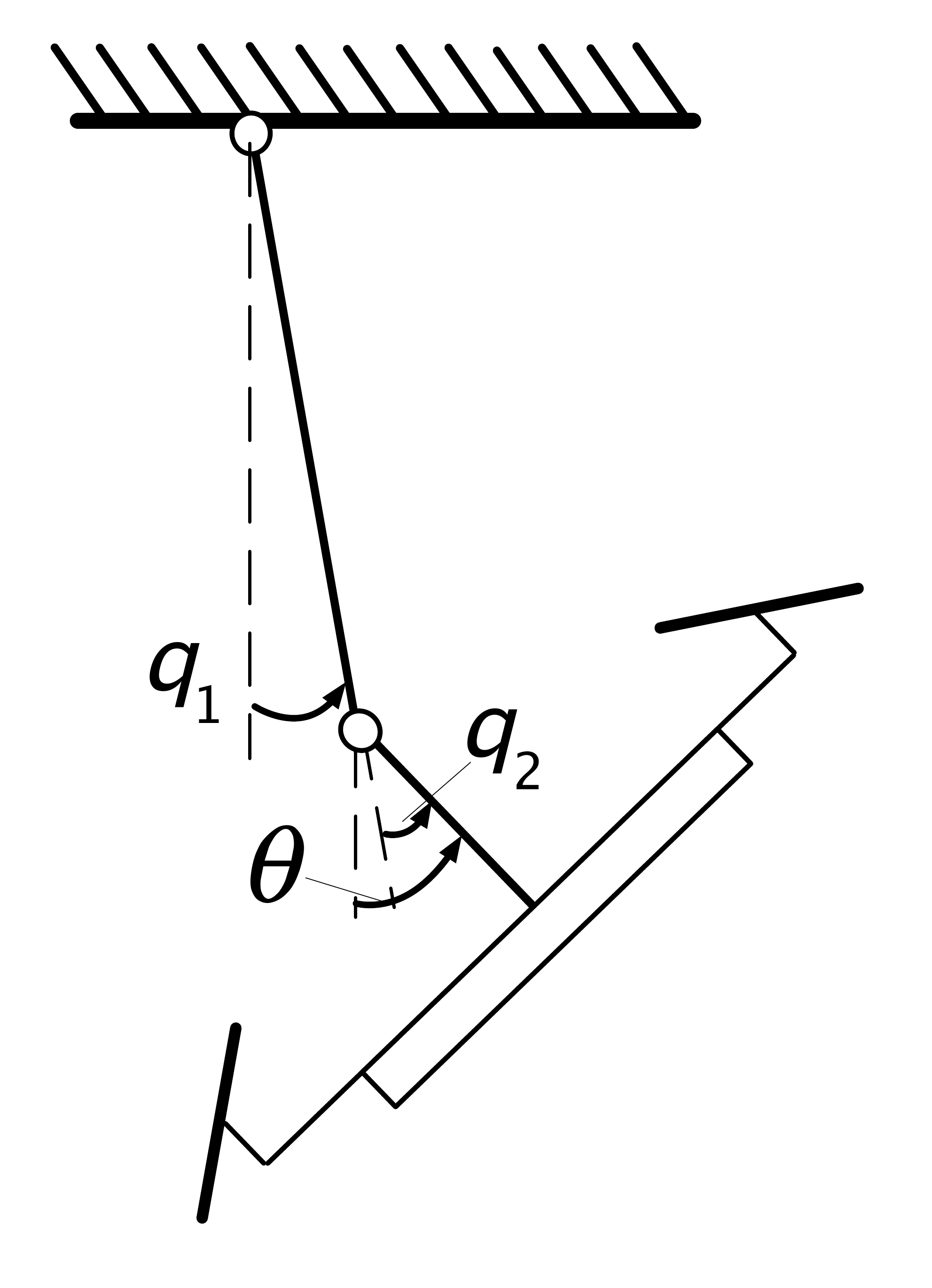}
		\caption{Angles definition}
		\label{fig:real_motion}
	\end{subfigure}
	\begin{subfigure}[]{0.47\linewidth}
		\centering
		\includegraphics[width=1\linewidth]{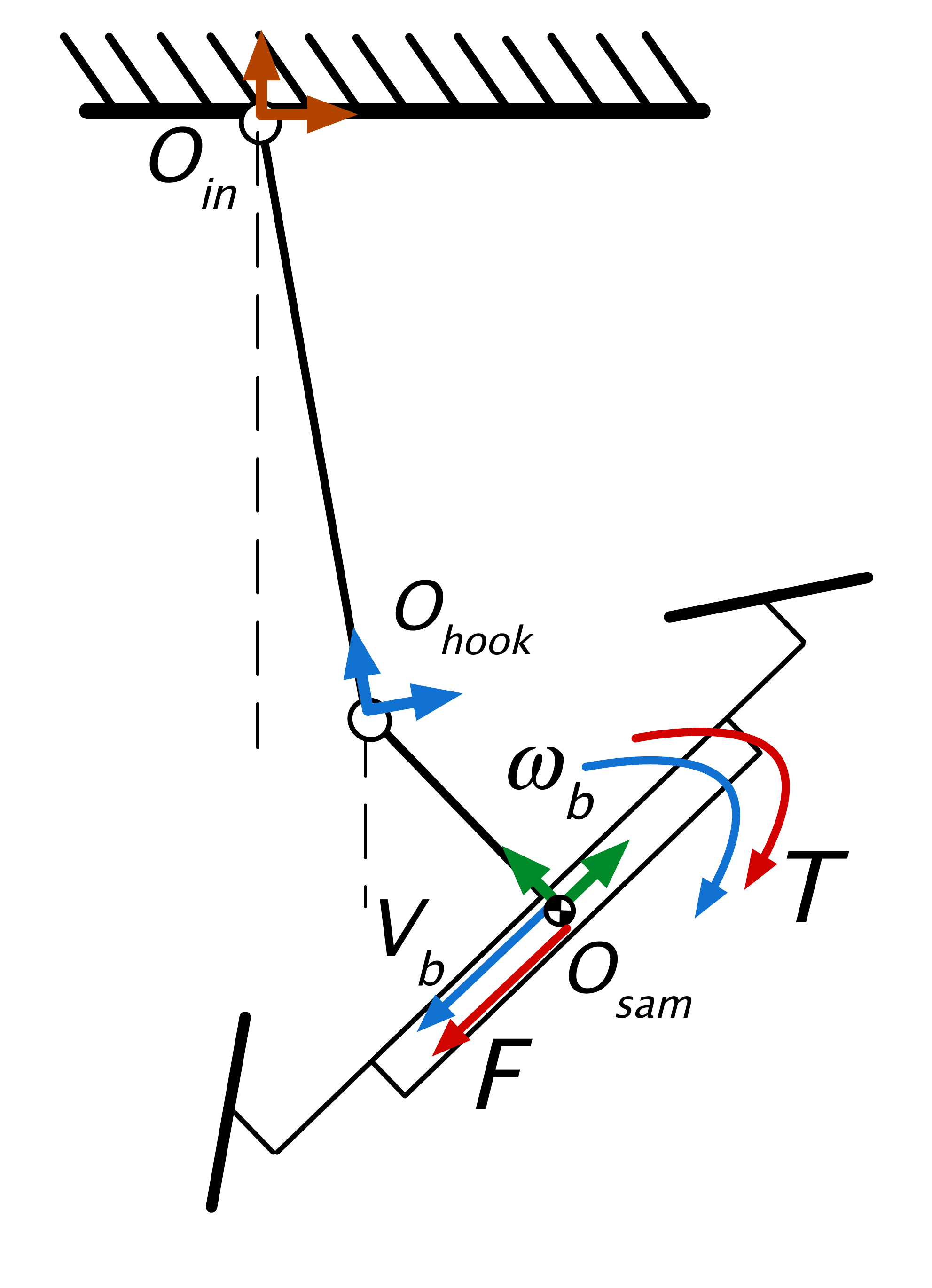}
		\caption{Body twist and wrench}
		\label{fig:real_motion2}
	\end{subfigure}
	%    \hfill
	\caption{Planar SAM.}
	\label{fig:planecase}
\end{figure}

Therefore, we can eliminate yaw- and manipulator-related variables in (\ref{eq:dynamics_model_full}), which then reduces to
\begin{align}
\label{eq:dynamics_model}
\tilde \bM (\tilde\bq) \ddot{\tilde \bq} + \tilde \bC (\tilde \bq, \dot{\tilde \bq}) \dot{\tilde \bq} +\tilde \bg (\tilde \bq) = \bm{\tilde \bJ^T \tilde \bu} + \bm{d}, 
\end{align}
where $\tilde{\bm{q}} = [\bm{\tilde{q}_1}^T \; \bm{\tilde{q}_2}^T]^T$ is a configuration (\ref{eq:config_orig}) without variables related to yaw and manipulator; $\tilde \bM$, $\tilde \bC$, $\tilde \bg$, and $\tilde \bJ$ represent components of remaining dynamics of (\ref{eq:dynamics_model_full}). Motion of manipulator causes an uncertain disturbance $\bm{d}$ that we further omit and treat as a source of oscillations that have to be dampened by controller. 

If we restrict the pendulum motion in the plane (see Fig. \ref{fig:planecase}), we can further simplify the model as follows:
\begin{align}
\label{eq:dynamics_model_planar}
\bar \bM (\bar\bq) \ddot{\bar \bq} + \bar \bC (\bar \bq, \dot{\bar \bq}) \dot{\bar \bq} +\bar \bg (\bar \bq) = \bar{\bJ}^T \bar{\bu},
\end{align}
where, $\bar{\bm{q}} = [q_1 \; q_2]^T$ corresponds to angles of first and second links, and $\bar{\bu} = [F \; T]^T$ is the control wrench, see Fig. \ref{fig:planecase}. The IMU measurement is then $\theta=q_1+q_2$ for the planar case. The body twist $\bV=[v_b \; w_b]^T$ and configuration $\dot{\bar{\bq}}$ are related by the Jacobian matrix:
\begin{align}
    \label{eq:Jacobian_bar}
    \underbrace{
    \left(
    \begin{array}{c}
    v_b \\
    w_b
    \end{array}
    \right)}_{=\bV}
    =
    \underbrace{\left[
    \begin{array}{cc}
     l_1 \cos{q_2}+l_2& l_2 \\
     1 & 1
    \end{array}
    \right]}_{=\bar{\bJ}}
    \underbrace{\left(
    \begin{array}{c}
    \dot{q}_1 \\
    \dot{q}_2
    \end{array}
    \right)}_{=\dot{\bar{\bq}}}.
\end{align}
In a certain range of $q_2$ ($q_2 < 90^{\circ}$), the Jacobian matrix $\bar{\bJ}$ is invertible. 

In the later section, we design a controller using the simplified planar model, while stability analysis is performed for the original system model (\ref{eq:dynamics_model}).

\subsection{Behavior of planar double pendulum}
\label{sec:understanding}

For later convenience, we present some insights on the behavior of planar double pendulum. Basically, any double pendulum system contains two oscillation motions \cite{magnus1965vibrations}: the first component with low frequency, $\nu_{slow}$, is modulating the second one with high frequency, $\nu_{fast}$. Although the solution for joint angles $q_1$ and $q_2$ contains both high and low frequency motions, in the system with parameter relation such as $l_1>l_2$ and $m_1<m_2$, roughly speaking, $q_1$ is dominated by slow motion and $q_2$ by fast motion.

Fast and slow frequencies of the planar double pendulum system can be calculated using the following:
\begin{align}
\label{eq:freq}
\nu_{fast,slow}^2=\frac{g m_{12}}{8 \pi^2 m_1 l_1 l_2}\bigg((l_{12})\pm  \sqrt{l_{12}^2-\frac{4m_1l_1l_2}{m_{12}}} \; \bigg).
\end{align}
Here, $m_1$ is a weight of the hook, $m_2$ is the weight of the platform, $m_{12} = m_1+m_2$, and $l_{12}=l_1+l_2$.
%Namely, $\dot{q}_1(t) \simeq C_{11} \dot{q}^{slow}$ and $\dot{q}_2(t) \simeq C_{22}\dot{q}^{fast}$ when $l_1 \gg l_2$.

%It is worth mentioning that, in the case of small angles, the $\dot {q}_1$ has dominant slow term, and $\dot {q}_2$ has as dominant the fast term.

%Frequencies of the modes corresponding to the slow and fast motions can be measured or precalculated, e.g., for the planar double pendulum system, frequencies can be expressed as:
%\begin{align}
%\nu_{fast,slow}^2=\frac{g}{8 \pi^2 m_1 l_1 l_2}\bigg((m_1+m_2)(l_1+l_2)\pm %\nonumber \\ \sqrt{(m_1+m_2)[(m_1+m_2)(l_1+l_2)^2-4m_1l_1l_2]} \; \bigg).
%%\nu_{1,2}^2=\frac{g}{2m l_2}\left((m+1)(1+l)\pm  %\sqrt{(m+1)[(m+1)(1+l)^2-4m l]} \; \right)
%\end{align}

\subsection{Control design}

In the planar model in (\ref{eq:dynamics_model_planar}), we first apply coordinate transformation from $\bar{\bq}$ to $\bV$ using (\ref{eq:Jacobian_bar}). Then, we obtain
\begin{align}
\label{eq:dynamics_model_in_twist}
\bm{\Lambda} (\bar\bq) \dot{\bV} + \bm{\Gamma} (\bar \bq, \dot{\bar \bq}) \bV +\bm{\zeta} (\bar \bq) =  \bar \bu,
\end{align}
where $\bm{\Lambda}, \bm{\Gamma}, \bm{\zeta}$ represent inertia, Coriolis/centrifugal, gravity in the new coordinate system. This coordinate transformation is valid since the Jacobian matrix $\bar{\bJ}$ is invertible. From (\ref{eq:dynamics_model_in_twist}), damping can be artificially injected by simply letting
\begin{align}
\label{eq:solution}
\bar{\bu}_{des} =
\begin{bmatrix}
F\\
T
\end{bmatrix}
= 
\begin{bmatrix}
- K_v v_b \\
- K_w w_b
\end{bmatrix},
\end{align}
where $K_v,K_w$ are positive control gains. Using (\ref{eq:solution}), the control goal addressed earlier can be achieved.

However, this control law cannot be directly applied because we have no measurements for $v_b$, whereas $w_b$ is directly obtained by IMU sensor. Assuming small angle for $q_2$ (hence $\cos(q_2) \simeq 1$), from (\ref{eq:Jacobian_bar}), $F$ can be approximated as
\begin{align}
    \label{eq:v_simplified_1}
    F = -K_v(l_1 \dot{q}_1 + l_2 {w_b}).
\end{align}

As addressed in Section \ref{sec:understanding}, $\dot{q}_1$ is dominated by slow oscillation motion with a low-frequency mode while $w_b$ contains both modes. Therefore, we can extract $\dot{q}_1$ from $w_b=\dot{\theta}$ by taking low-pass filter, and then  (\ref{eq:solution}) can be rewritten as
\begin{align}
    \label{eq:control_F_simplified_2}
%    F = -K_v (l_1 w_b^{lp} + l_2 w_b),
\bar{\bu} =
\begin{bmatrix}
F\\
T
\end{bmatrix}
= 
\begin{bmatrix}
-K_v (l_1 w_b^{lp} + l_2 w_b) \\
- K_w w_b
\end{bmatrix},
\end{align}
where
\begin{align}
    w_b^{lp} = \frac{1}{\tau s+1}w_b
\end{align}
is the low-pass filter with the time constant $\tau$.

We extend the presented control law to the original system (\ref{eq:dynamics_model}) as follows
\begin{align}
\label{eq:solution_for_3d}
\tilde{\bu} =
\begin{bmatrix}
\bF\\ \bT
\end{bmatrix}
= 
\begin{bmatrix}
- \bm{K}_v (l_1 \bm{w}_b^{lp} + l_2 \bm{w}_b)\\ 
- \bm{K}_w \bm{w}_b.
\end{bmatrix} 
\end{align}
In our control design, the cut-off frequency of the low-pass filter is selected in the middle between slow and high frequency modes, i.e., 
\begin{align}
\label{eq:cutoff}
\nu_{cutoff}=\frac{\nu_{slow}+\nu_{fast}}{2}.
\end{align}

%The time constant for the low-pass filter, can be calculated as the inverse of cut-off frequency, which can be be selected in the middle of frequency range $\frac{\nu_{slow}+\nu_{fast}}{2}$.
%\begin{align}
%\label{eq:cutoff}
%\nu_{cutoff}=\frac{\nu_{slow}+\nu_{fast}}{2}.
%\end{align}

\subsection{Closed-loop stability}
\label{sec:stability}
Since the double pendulum system with damping in joints is asymptotically stable by nature, we investigated the stability of our controller in a simulator. We report that for the $l_1 = 4..10 [m]$ while changing the initial angles of the passive joints from 2 till 45 degrees in arbitrary configuration with a step of 7 degrees, the closed-loop system was always stable. Moreover, initial angular velocities at these joints were varied from 0 to 1 $\frac{rad}{s}$ with a step of 0.5. Performed analysis also confirms that the proposed controller is robust against model uncertainties. 
%One limitation of presented stability analysis is the fact that it is valid only locally. Nevertheless, in our experience, this controller works also with initial conditions far from operating point. To provide some evidence, we present numerical simulation in Section \ref{sec:validation}. 

\subsection{Gain tuning rule}

Since the system is stable for (almost) any choice of parameters and gains, it is important to seek the best gain in some sense. In particular, we seek for the control gains which minimize the following linear quadratic cost function
\begin{align}
\label{eq:costf}
J =
\int_0^t  \left( \bX(t)^T \bQ\bX(t) +
\bU(t)^T \bR \bU(t) \right) \; dt,
\end{align}
where $\bQ \geq \bzero$ penalizes the state $\bX$, and $\bR > \bzero$ penalizes the amount of control input $\bU$.

In this subsection, we again use the planar double pendulum for simplicity. Also, we use $\dot{\theta}$ (which is equivalent to $w_b$ in planar case) for the body angular velocity in this subsection just to have a nicer look. Let us first linearize  (\ref{eq:dynamics_model_planar}) around operating point as follows
\begin{align}
    \label{eq:linear_model}
    \dot{\bX} =& \bm{A} \bX + \bm{B}\bU,
\end{align} 
where $\bX = [q_1 \;\; \dot{q}_1 \;\; \theta \;\; \dot{\theta} \;\; \dot{\theta}^{lp}]^T$, with
\begin{align}
\bA =& 
\begin{pmatrix}
0 & 1 & 0 & 0 & 0 \\%& 0 \\
- \frac{gm_{12}}{m_1 l_1} & 0 & \frac{m_2 g}{m_1 l_1} & 0 & 0 \\%& \frac{K_w}{m_1 l_1 l_2} \\
0 & 0 & 0 & 1 & 0 \\% & 0 \\
\frac{gm_{12}}{m_1 l_2} & 0 & - \frac{gm_{12}}{m_1 l_2} & 0 & 0\\% &  - \frac{K_w m_{12}}{m_1 m_2 l_2^2}\\
0 & 0 & 0 & \frac{1}{\tau} & -\frac{1}{\tau}
\end{pmatrix}, \\
\bm{B} =& 
\begin{pmatrix}
0 & 0  \\%& 0 \\
0 & -\frac{1}{m_1 l_1 l_2} \\%& \frac{K_w}{m_1 l_1 l_2} \\
0 & 0 \\% & 0 \\
\frac{1}{m_2 l_2} & \frac{m_{12}}{m_1 m_2 l_2^2}\\% &  - \frac{K_w m_{12}}{m_1 m_2 l_2^2}\\
0 & 0
\end{pmatrix}.
\end{align}

The last row of matrix $\bm{A}$ corresponds to the dynamics of the low-pass filter:
\begin{align}
    \tau \ddot{\theta}^{lp} + \dot{\theta}^{lp} = \dot{\theta}.
\end{align}

To make the control input (\ref{eq:control_F_simplified_2}) have output feedback form, we define output as
\begin{align}
    \label{eq:measurement}
    \bY = \bm{C} \bX,
\end{align}
where
\begin{align}
    \bm{C}=
\begin{pmatrix}
    0 & 0 & 0 & l_2 & l_1 \\
    0 & 0 & 0 & 1 & 0
\end{pmatrix}. 
\end{align}

Consequently, the control input $\bU$ is expressed as the output feedback form:
\begin{align}
\label{eq:stab_control}
    \bU = - \bF \bY = - 
    \left[
    \begin{array}{cc}
    K_v & 0  \\
    0 & K_w 
    \end{array}
    \right]\bY.    
\end{align}

For our system described by (\ref{eq:linear_model})-(\ref{eq:measurement}), we applied the output feedback LQR technique which can be formulated using linear matrix inequalities (LMIs).

\begin{thm} 
Let us consider the system (\ref{eq:linear_model}) with the output (\ref{eq:measurement}). There exists an optimal controller in the form of (\ref{eq:stab_control}) which minimizes the cost function (\ref{eq:costf}), if the following problem has a solution for the given matrix $\bXi>\bzero$ and weighting matrices $\bm{Q} \leq \bzero$, $\bm{R}>\bzero$:
\begin{align}
\min_{\bF,\bP} trace(\bP),
\end{align}
subject to LMIs:
\begin{align}
    \bM \leq 0, \;\; \bP>0, \nonumber
\end{align}
where
\begin{align}
    \bM =
    \left[
    \begin{array}{cc}
    \bA^T \bP + \bP\bA + \bQ + \bm{H} & \bG^T \\ 
    \bG & -\bR^{-1}
    \end{array}
    \right],
\end{align}
with
\begin{align}
    \bG =& \bF\bC - \bR^{-1}\bm{B}^{T} \bP,  \\
\nonumber    \bm{H} =  & -(\bXi \bm{B})\bR^{-1}(\bm{B}^T\bP) -(\bP \bm{B})\bR^{-1}(\bm{B}^T \bXi)  \\ 
    &+(\bXi \bm{B})\bR^{-1}(\bm{B}^T\bXi). 
\end{align} 
\end{thm}

%MOSEK package \cite{andersen2000mosek} 
To solve the given LMI problem, we used oflqr library \cite{ilka2018matlab} with the LMI solver in YALMIP \cite{lofberg2004yalmip}. In our control design, we selected $\bm{Q}$ and $\bm{R}$ matrices as follows:
\begin{align}
\label{eq:QandR}
    \bm{Q}=diag\{0,10,0,1,0\} \; \text{and} \; \bm{R}=\sigma \cdot diag\{1,10\}.
\end{align}

Since our control goal is to dampen the oscillations, we penalized only $\dot{q}_1$ and $\dot{\theta}$. Moreover, we applied stronger control action on slow oscillation mode which might be more critical when performing manipulation tasks in a real industrial scenario. For this reason, we penalized $\dot{q}_1$ more than $\dot{\theta}$ in $\bm{Q}$, and allowed more control input for $F$ in $\bm{R}$ design.

\begin{figure}
	    \centering
	    \includegraphics[scale=0.45]{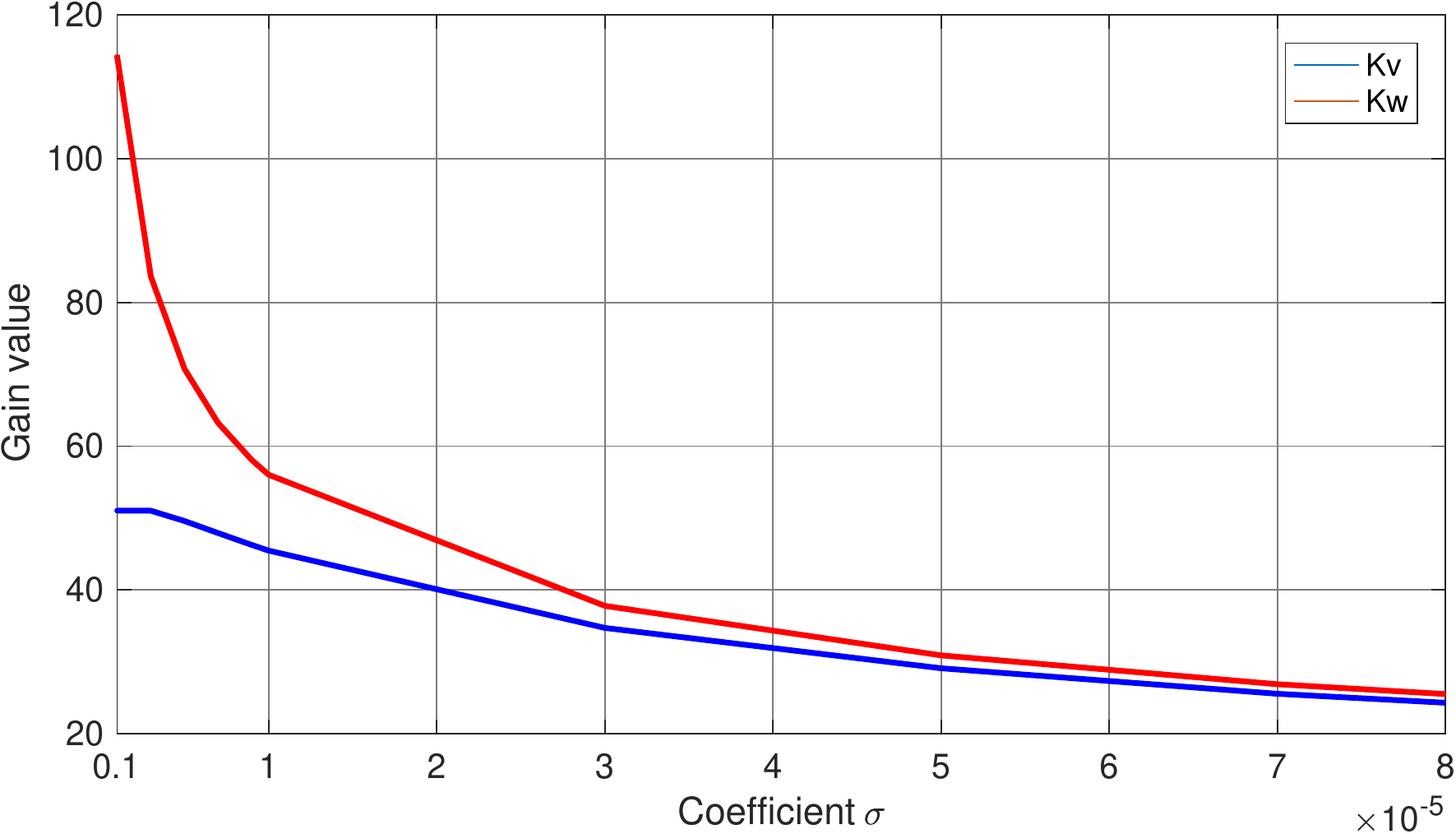}
	    \caption{Relation between optimal control gains $K_v, K_w$ and parameter $\sigma$.}
	\label{fig:optimization}
\end{figure}

In (\ref{eq:QandR}), $\sigma$ is a new parameter that allows us to investigate optimal control gains over admissible control inputs; note that smaller $\bm{R}$ implies larger control input. Therefore, we solved the optimization problems with varying $\sigma$: from $1e^{-6}$ to $8e{-5}$, as shown in Fig. \ref{fig:optimization}. Depending on the designer's choice (balance between oscillation damping and power consumption), we can select one good combination of gains.%It can be seen, that ratio of $K_w$ to $K_v$ is decreasing from 2 to 1 while $\sigma$ is growing in the considered range.

\begin{figure}[t]
	\centering
	\includegraphics[scale=0.33]{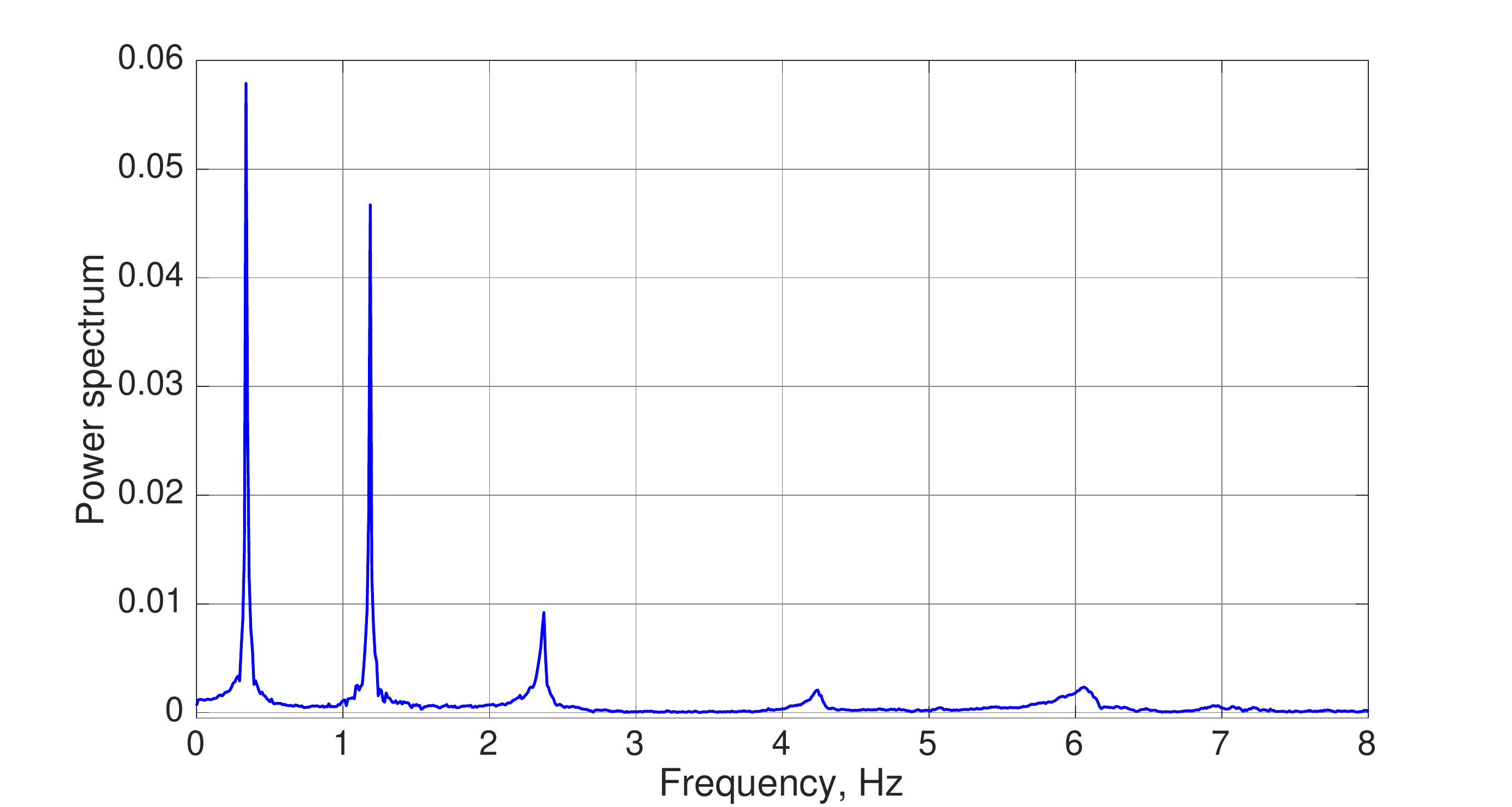}
	\caption{Power spectrum of real system.}
	\label{fig:power}
\end{figure}

Optimization was conducted using parameters measured in the real system:
\begin{align}
\label{eq:parameters}
m_1 = 18.5 \; [kg],\;\; m_2 = 55\; [kg],\;\; l_1 = 6\; [m],\;\; l_2 = 2.2\; [m]  
\end{align}
As a time constant for low-pass filter, cut-off frequency $\nu_{cutoff} = 0.76 [Hz]$ was calculated based on (\ref{eq:cutoff}). Oscillation mode frequencies $\nu_{slow}$ and $\nu_{fast}$ were obtained from the real hardware, as shown in Fig. \ref{fig:power}.

\begin{figure}[t]
	\centering
		\includegraphics[scale=0.45]{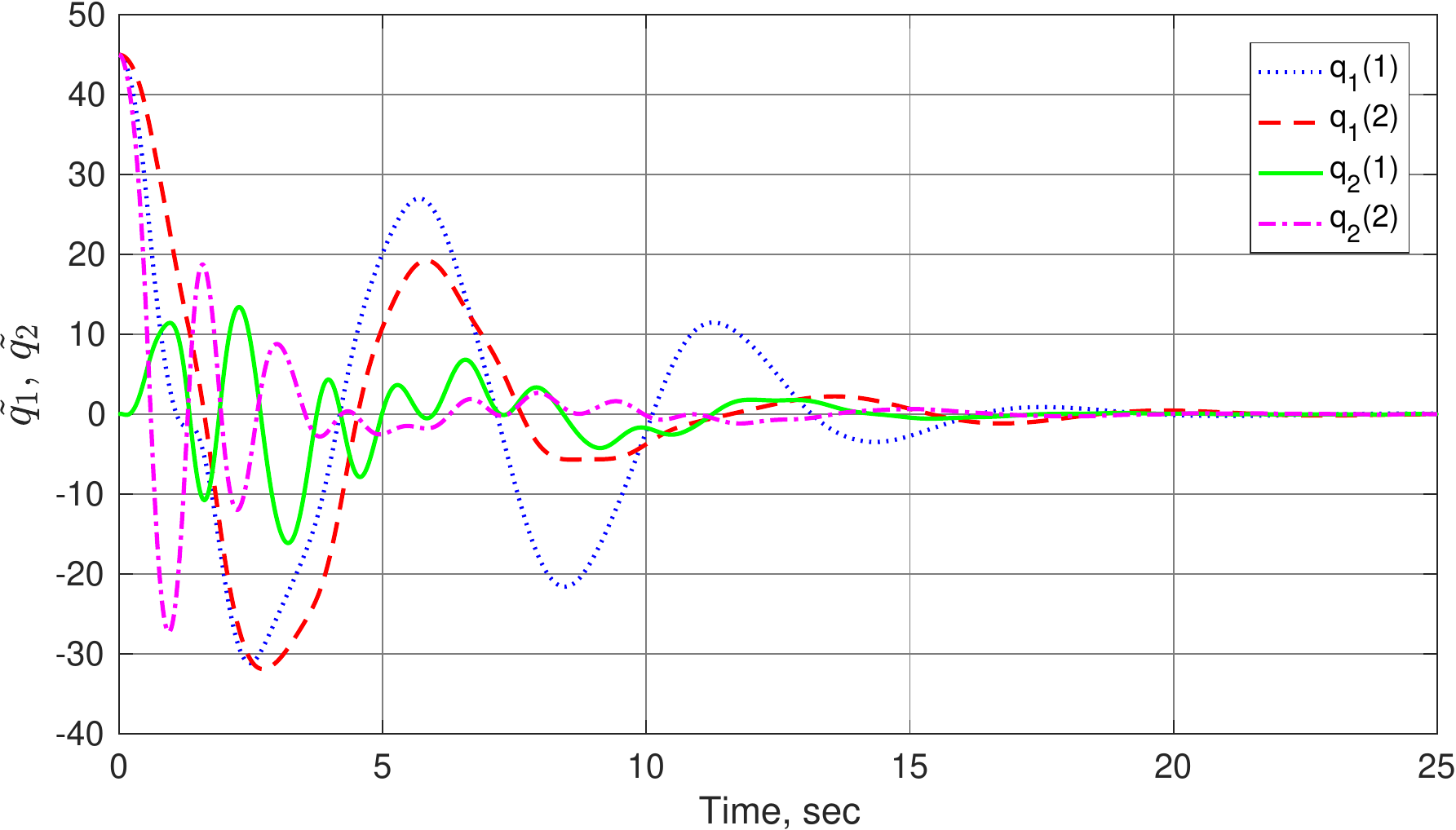}
		\caption{State convergence with extreme initial conditions.}
		\label{fig:stabil}
	\end{figure}
	
\begin{figure}[t]
	\centering
		\includegraphics[scale=0.45]{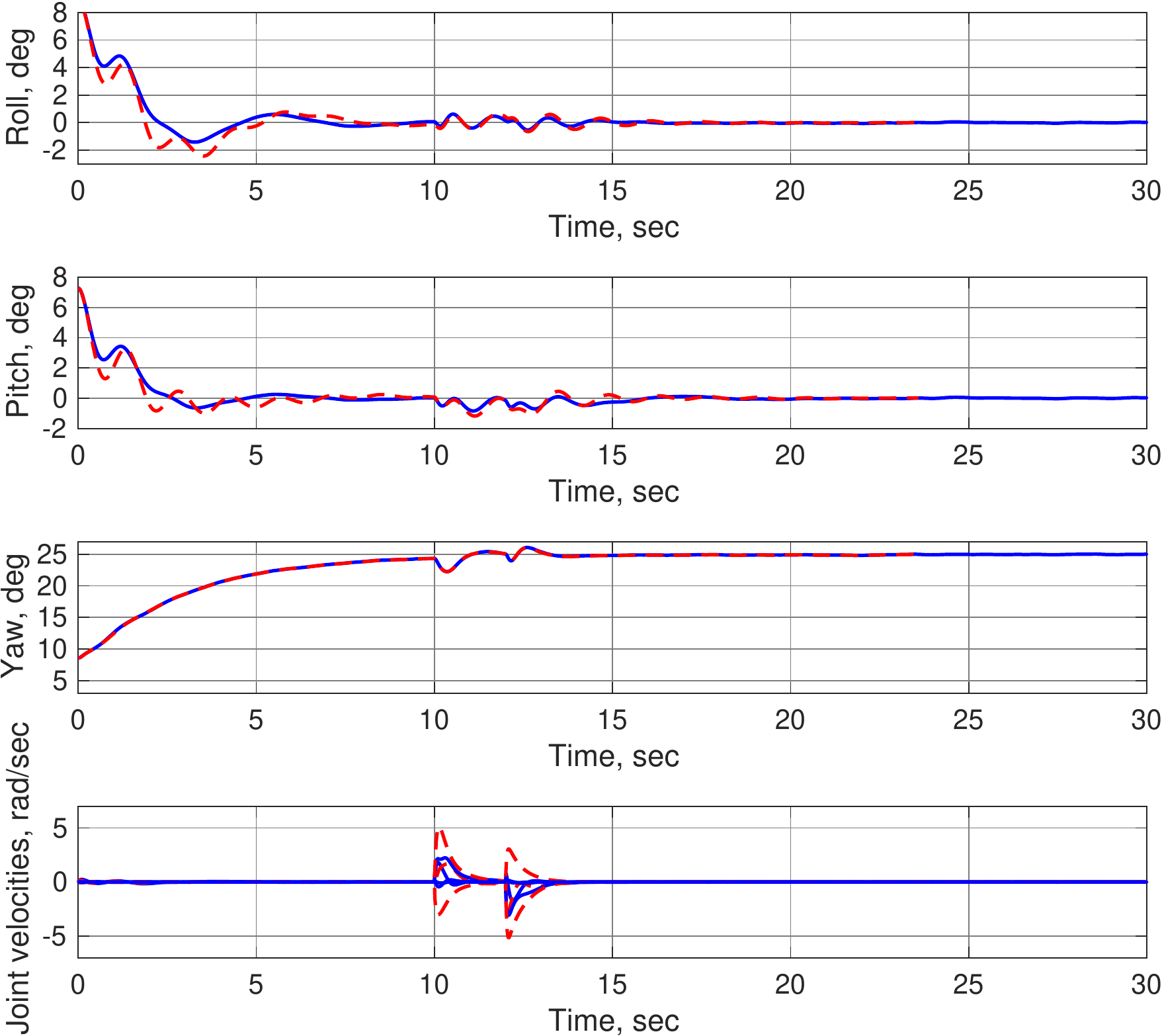}
		\caption{Comparison of the proposed (blue) and ideal (red) controllers.}
		\label{fig:stabil3}
\end{figure}

\section{Validation}
\label{sec:validation}

In order to validate the proposed controller, numerical simulation and experiments were conducted. For both cases, to control the robotic arm and to perform oscillation damping while keeping constant yaw, the decentralized control approach was applied as we addressed in Section \ref{sec:reduced_model}.

\subsection{Validation of double pendulum model}

To see the validity of the double pendulum model, we investigated oscillation modes of the real hardware during the free motion. To this end, we lifted up the SAM and then released. We applied fast Fourier transform to obtain a power spectrum. As shown in Fig. \ref{fig:power}, there exist two dominant frequency modes, and therefore, it is reasonable to model the system as a double pendulum.

%It is worth mentioning that the presence of two dominant modes in the spectrum of the real system confirms that the system can be modeled as a double pendulum for investigation. 

\subsection{Numerical simulations}
Numerical simulation was conducted based on \cite{garofalo2013closed} which proposed an algorithm that computes dynamic parameters efficiently. The SAM platform was modeled as a spherical double pendulum. In this simulation, all parameters were chosen to be as close as possible to the real setup. To this end, in addition to the model parameters given in (\ref{eq:parameters}), gyro noise density $0.009 \; [\circ/s/\sqrt{Hz}]$ was taken from calibration certificate provided by the manufacturer of IMU. However, influence of unmodeled dynamics (e.g., weight of the link), airflow, and actuator dynamics is not considered in the simulation. Performance under all uncertainties will be validated through experiments. %In the simulation, the selected optimal control gains were $K_v = 52, K_w =63$ with $\sigma = 5 e^{-6}$.

\begin{figure}[t]
		\centering
		\includegraphics[scale=0.055]{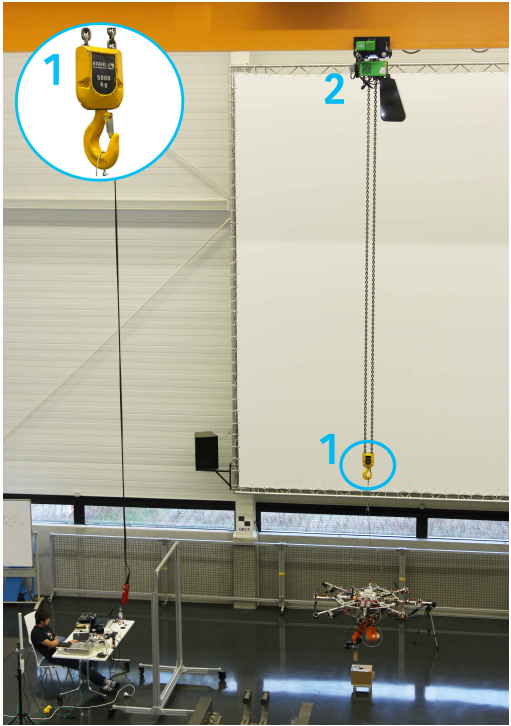}
		\caption{Experimental setup: 1 is hook, 2 is winch.}
		\label{fig:exp_setup}
\end{figure}

In this subsection, we present two simulation studies. First, we validate stability of the proposed controller.  As addressed in Section \ref{sec:stability}, we investigated stability for large variations of initial conditions including those which are far from normal operating range. In our experience, the SAM was never excited more than $5^\circ$ for roll/pitch. Nevertheless, we set $45^\circ$ for all initial angle values as an extreme case. As shown in Fig. \ref{fig:stabil}, all system states converged to the equilibrium point, which indicates asymptotic stability. 

Second, we compare the proposed control law (\ref{eq:solution_for_3d}), and the ideal controller
\begin{align}
    \bu_{des} = 
\begin{bmatrix}
\bm{F}\\ \bm{T}
\end{bmatrix}
= 
\begin{bmatrix}
- K_v \bm{v}_b \\ - K_w \bw_b
\end{bmatrix},
\end{align}
which is the 3D version of (\ref{eq:solution}). Recall that we proposed (\ref{eq:solution_for_3d}) because $\bm{v}_b$ is not measurable for the real system. As shown in Fig. \ref{fig:stabil3}, overall shapes of resulting behavior were quite similar for both controllers. Actually, the proposed controller resulted in less oscillations compared to the ideal one due to effect of filters. At $t=10$, the robotic arm was commanded to cause jerky motions to apply some disturbances on the SAM. The simulation validates that the proposed control could dissipate the oscillation caused by the disturbances.

%In this simulation, system starts motion with some initial conditions, while after stabilization robotic arm caused some additional disturbances by performing jerky motion with high joint velocities. 

%Numerical simulation was conducted based on \cite{garofalo2013closed} which proposed an algorithm that computes dynamic parameters efficiently. The SAM platform was modeled as a double spherical double pendulum. In this simulation, all parameters were chosen to be as close as possible to the real setup. To this end, in addition to the model parameters given in (\ref{eq:parameters}), gyro noise density $0.009 \; [\circ/s/\sqrt{Hz}]$ was taken from calibration certificate \mj{provided by the manufacturer?}. However, influences of unmodeled dynamics (e.g., weight of the link), airflow, and actuator dynamics are not considered in simulation. Validation under all uncertainties will be validated  through experiments. 

%It is worth mentioning that the presence of two dominant modes in the spectrum of the real system confirms that the system can be modeled as a double pendulum for investigation. 

\subsection{Experimental validation}

To corroborate simulation results, experimental validation was carried out using the overhead crane. Our experimental setup is shown in Fig. \ref{fig:exp_setup}. Blue digits mark the elements (winch and hook) which correspond to the two passive spherical joints. Hook has a passive DoF around the yaw, i.e., hook itself can rotate around the hook base. The chain of the winch can twist around a vertical axis.

\begin{figure}[t]
	\centering
	\includegraphics[trim={35 22 0.1 26},clip,scale=0.49]{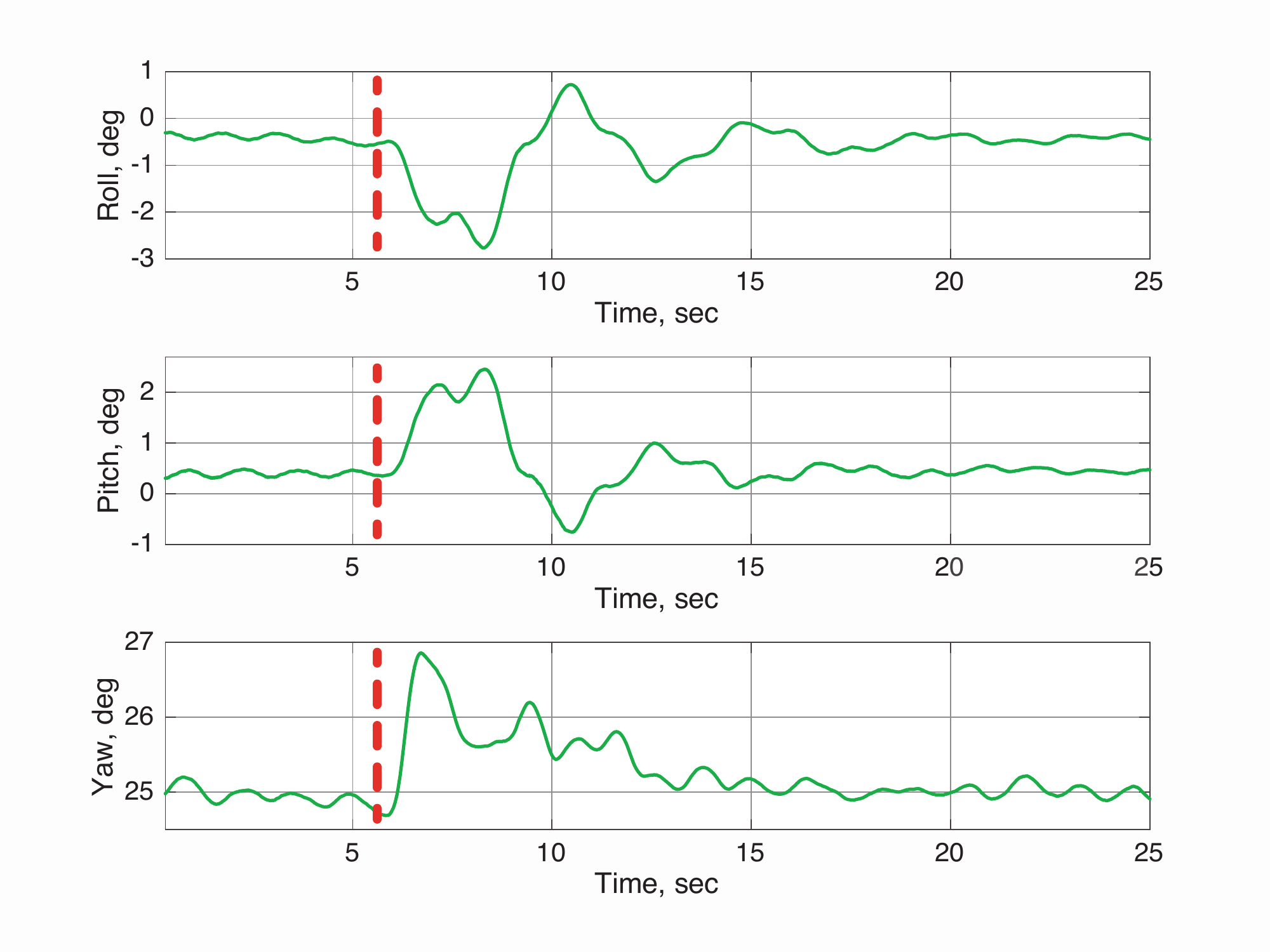}
	\caption{Experimental results. Red dashed line marks the time at which external disturbances were applied to the system.}
	\label{fig:exp_res}
\end{figure}

The selected optimal control gains were $K_v = 48, K_w = 70$ with $\sigma = 5 e^{-6}$. In the experiment, human applied external disturbance to excite the SAM, while the controller was trying to dampen any oscillations. As shown in Fig. \ref{fig:exp_res}, the controller quickly damped out the oscillation within 6 seconds. Please see also the video attachment in which we have compared the convergence rate of the passive system and controlled system under the effect of different disturbances, e.g., external perturbations, moving suspension point, and jerky motion of the robotic arm.

%the convergence time for the angles measured by onboard IMU is almost equal to the simulated one.

%\section{Conclusions and Future work}

\section{Summary and Concluding Remarks}
\label{sec:end}
In this paper, the oscillation damping control approach for the SAM platform was designed, applied, and validated in simulation and experimental studies. The system in operation can be considered as a spherical double pendulum, which can be controlled only indirectly by generating a damping wrench at the tip of the second link. Due to the absence of the state measurements, damping wrench was generated by proposed controller using only onboard IMU without any model parameters. Additionally, we found the optimal control gains which minimize the linear quadratic cost function, so the resulting controller dissipates the oscillation with desired balance between performance and power consumption. Moreover, by virtue of the optimal gains, we can easily tune the gains for different operating conditions.
%The proposed controller is a simple yet robust solution to this problem, because it does not require any model parameters and because it utilizes only a single onboard IMU sensor. 

Although local stability could be shown theoretically, we presented some evidence from which stability in reasonably large operating range could be expected. Performance of the controller can be further improved by considering the motion of the robotic arm as a part of the system and by fusing additional sensor information, e.g., GPS RTK or 3D-vision. In the future, we plan to design a controller that adapts to varying link length (which is controllable by the crane) and battery status (for instance, more weights on the control input when the battery is low).

\bibliographystyle{IEEEtran.bst}
\newpage

\bibliography{mybib.bib}

\end{document}